\documentclass[10pt,twocolumn,letterpaper]{article}

\usepackage{cvpr}
\usepackage{times}
\usepackage{epsfig}
\usepackage{graphicx}
\usepackage{amsmath}
\usepackage{amssymb}
\usepackage{makecell}
\usepackage{subcaption}
\usepackage{mathtools, nccmath}
\usepackage{gensymb}
\usepackage{dblfloatfix}

\DeclarePairedDelimiter{\nint}\lfloor\rceil

\usepackage[pagebackref=true,breaklinks=true,letterpaper=true,colorlinks,bookmarks=false]{hyperref}

\cvprfinalcopy 

\def\cvprPaperID{31} 

\ifcvprfinal\fi

\begin{document}

\title{GolfDB: A Video Database for Golf Swing Sequencing}

\author{William McNally\quad\quad Kanav Vats\quad\quad Tyler Pinto\quad\quad Chris Dulhanty\\John McPhee\quad\quad Alexander Wong\\
Systems Design Engineering, University of Waterloo\\
{\tt\small \{wmcnally, k2vats, tyler.pinto, chris.dulhanty, mcphee, a28wong\}@uwaterloo.ca}
}

\maketitle

\begin{abstract}
\vspace{-5pt}
The golf swing is a complex movement requiring considerable full-body coordination to execute proficiently. As such, it is the subject of frequent scrutiny and extensive biomechanical analyses. In this paper, we introduce the notion of golf swing sequencing for detecting key events in the golf swing and facilitating golf swing analysis. To enable consistent evaluation of golf swing sequencing performance, we also introduce the benchmark database \textbf{GolfDB},\footnote{Available at \url{https://github.com/wmcnally/GolfDB}} consisting of 1400 high-quality golf swing videos, each labeled with event frames, bounding box, player name and sex, club type, and view type. Furthermore, to act as a reference baseline for evaluating golf swing sequencing performance on GolfDB, we propose a lightweight deep neural network called SwingNet, which possesses a hybrid deep convolutional and recurrent neural network architecture. SwingNet correctly detects eight golf swing events at an average rate of 76.1\%, and six out of eight events at a rate of 91.8\%.  In line with the proposed baseline SwingNet, we advocate the use of computationally efficient models in future research to promote in-the-field analysis via deployment on readily-available mobile devices. 
\end{abstract}

\vspace{-10pt}
\section{Introduction}
It is estimated that golf is played by 80 million people worldwide \cite{hsbc2020golf}. The sport is most popular in North America, where 54\% of the world's golf facilities reside~\cite{ra2017golf}. In the United States, the total economic impact of the golf industry is estimated to be \$191.9 billion~\cite{golf2020}. In Canada, golf has had the highest participation rate of any sport since 1998~\cite{gc2013sport}. It would be reasonably contended that many golfers are drawn to the sport through the gratification of continuous improvement. The golf swing is a complex full-body movement requiring considerable coordination. As such, it can take years of practice and instruction to develop a repeatable and reliable golf swing. For this reason, golfers routinely scrutinize their golf swing and make frequent adjustments to their golf swing mechanics.

Several methods exist for analyzing golf swings. In scientific studies, researchers distill golf swing insights using optical cameras to track reflective markers placed on the golfer~\cite{m1998hip, lephart2007eight, myers2008role, chu2010relationship}. Often, these insights relate to kinematic variables at various \textit{events} in the golf swing. For example, in \cite{myers2008role} it was found that torso--pelvic separation at the top of the swing, referred to as the X-factor in the golf community, is strongly correlated with ball speed. Similarly, in \cite{chu2010relationship} it was found that positive lateral bending of the trunk at impact (away from the target) was also correlated with ball speed, as it potentially promotes the upward angle of the clubhead path and more efficient impact dynamics~\cite{mcnally2018dynamic}. Yet, examining a golf swing using motion capture requires special equipment and is very time consuming, making it impractical for the everyday golfer. 

\begin{figure}[t]
    \centering
    \includegraphics[width=1.0\linewidth]{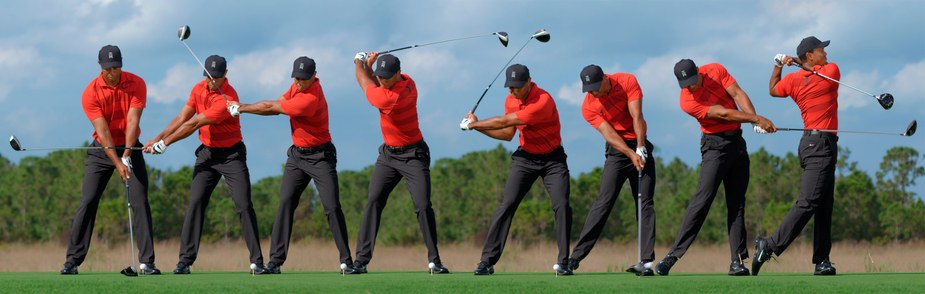}
    \includegraphics[width=1.0\linewidth]{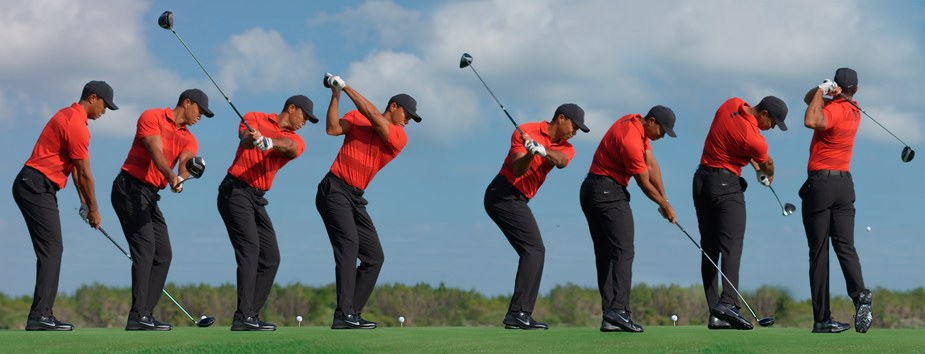}
    \caption{Eight events in a golf swing sequence. Top: face-on view. Bottom: down-the-line view. The names of the events from left to right are \textit{Address}, \textit{Toe-up}, \textit{Mid-backswing}, \textit{Top}, \textit{Mid-downswing}, \textit{Impact}, \textit{Mid-follow-through}, and \textit{Finish}. Images used with kind permission from Golf Digest~\cite{golfdigest}.}
    \label{fig:teaser}
    \vspace{-15pt}
\end{figure}

Traditionally, professional golf instructors provide instant feedback to amateurs using the naked eye. Still, the underlying problem is not always immediately apparent due to the speedy nature of the golf swing. Consequently, slow-motion video has become a popular medium for dissecting the intricacies of the golf swing~\cite{guadagnoli2002efficacy}. Moreover, slow-motion video is readily available to the common golfer using the advanced optical cameras in today's mobile devices, which are capable of recording high-definition (HD) video at upwards of 240 frames per second (fps). Given a slow-motion recording of a golf swing, a golfer or golf instructor may scrub through the video to analyze the subject's biomechanics at various key events. These events comprise a golf swing \textit{sequence}~\cite{golfdigest}. For example, in the golf swing sequence of Tiger Woods depicted in  Fig.~\ref{fig:teaser}, the X-Factor at the top of the swing and lateral bending of the trunk at impact, two strong indicators of a powerful golf swing, are easily identifiable in the face-on view. Still, scrubbing through a video to identify these events is time consuming and impractical because only one event can be viewed at a time.

In computer vision, deep convolutional neural networks (CNNs) have recently been shown to be highly proficient at video recognition tasks such as video representation~\cite{carreira2017quo}, action recognition~\cite{mcnally2019starnet}, temporal action detection~\cite{zhao2017temporal}, and spatio-temporal action localization~\cite{el2018real}. Following this line of research, CNNs adapted for video may be leveraged to facilitate golf swing analysis through the autonomous extraction of event frames in golf swing videos. To this end, we introduce \textbf{GolfDB}, a benchmark video database for the novel task of \textit{golf swing sequencing}. A total of 1400 HD golf swing videos of male and female professional golfers, comprising various native frame-rates and over 390k frames, were gathered from YouTube. Each video sample was manually annotated with eight event labels (event classes shown in Fig.~\ref{fig:teaser}). Furthermore, the dataset also contains bounding boxes and labels for club type (\textit{e.g.}, driver, iron, wedge), view type (face-on, down-the-line, or other), and player name and sex. With this supplemental data, GolfDB creates opportunities for future research relating to general recognition tasks in the sport of golf. Finally, we advocate mobile deployment by proposing a lightweight baseline network called \textbf{SwingNet} that correctly detects golf swing events at a rate of 76.1\% on GolfDB.

\section{Related Work}

\subsection{Computer Vision in Golf}
Arguably the most well-known use of computer vision in golf deals with the real-time tracing of ball flights in golf broadcasts~\cite{toptracer}. The technology uses difference images between consecutive frames and a ball selection algorithm to artificially trace the path of a moving ball. In a different light, radar vision is used in the TrackMan launch monitor to precisely track the 3D position of a golf ball in flight and measure its spin magnitude~\cite{trackman}. 

Computer vision algorithms have also been implemented for analyzing golf swings. Gehrig \textit{et al.}~\cite{gehrig2003visual} developed an algorithm that robustly fit a global swing trajectory model to club location hypotheses obtained from single frames. Fleet \textit{et al.}~\cite{urtasun2005monocular} proposed incorporating dynamic models with human body tracking, and Park \textit{et al.}~\cite{park2017accurate} developed a prototype system to investigate the feasibility of pose analysis in golf using depth cameras. In line with these research directions, GolfDB may be conveniently extended in the future to include various keypoint annotations to support human pose and golf club tracking. Moreover, automated golf swing sequencing using SwingNet is complementary to pose-based golf swing analysis. 

\subsection{Action Detection}

In the domain of action detection, there exist several sub-problems that correspond to increasingly complex tasks.

Action recognition is the highest-level task and corresponds to predicting a single action for a video. The first use of modern CNNs for action recognition was by Karpathy \textit{et al.} \cite{karpathy2014large}, wherein they investigated multiple methods of fusing temporal information from image sequences. Simonyan and Zisserman \cite{simonyan2014two} followed this work by incorporating a second stream of optical flow information to their CNN. A different approach to combine temporal information is to use 3D CNNs to perform convolution operations over an entire video volume. First implemented for action recognition by Baccouche \textit{et al.}\ \cite{baccouche2011sequential}, 3D CNNs showed state-of-the-art performance in the form of the C3D architecture on several benchmarks when trained on the Sports-1M dataset \cite{tran2015learning}. Combining two-stream networks and 3D CNNs, Carreira and Zisserman~\cite{carreira2017quo} created the I3D architecture.  Recurrent neural networks (RNNs) provide a different approach to combining temporal information. RNNs with long short-term memory (LSTM) cells are well suited to capture long-term temporal dependencies in data \cite{hochreiter1997long} and these networks were first used for action recognition by Donahue \textit{et al.}~\cite{donahue2015long} in the long-term recurrent convolutional network (LRCN), whereby features extracted from a 2D CNN were passed to an LSTM network. A similar method to Donahue \textit{et al.}\ is adopted in this work.

Temporal action detection is a mid-level task wherein the start and end frames of actions are predicted in untrimmed videos. Shou \textit{et al.}~\cite{shou2016temporal} proposed the Segment-CNN (S-CNN) in which they trained three networks based on the C3D architecture. In a different approach, Yeung \textit{et al.}~\cite{yeung2016end} built an RNN that took features from a CNN and utilized reinforcement learning to learn a policy for identifying the start and end of events, allowing for the observation of only a fraction of all video frames.


Spatio-temporal action localization is the lowest level and most complex task in action detection. Both the frame boundaries and the localized area within each frame corresponding to an action are predicted. Several works in this domain approach the problem by combining 3D CNNs with object detection models, such as in \cite{girdhar2018better}, where the I3D model is combined with Faster R-CNN, and in \cite{hou2017tube}, where the authors generalize the R-CNN from 2D image regions to 3D video \textit{tubes} to create Tube-CNN (T-CNN). 


\subsection{Event Spotting}
After asking Amazon Mechanical Turk workers to re-annotate the temporal extents of human actions in the Charades~\cite{sigurdsson2016hollywood} and MultiTHUMOS~\cite{yeung2018every} datasets, Sigurdsson \textit{et al.\ }\cite{sigurdsson2017actions} demonstrated that the temporal extents of actions in video are highly ambiguous; the average agreement in terms of temporal Intersection-over-Union (tIOU) was only 72.5\% and 58.7\%, respectively. This raised concerns over the inherent error associated with temporal action detection.

Considering the uncertainty surrounding the temporal extents of actions, Giancola \textit{et al.}~\cite{giancola2018soccernet} proposed the task of \textit{event spotting} within the context of soccer videos, arguing that in contrast to actions, events are anchored to a single time instance and are defined using a specific set of rules. As opposed to predicting the temporal bounds of an action, \textit{spotting} consists of finding the time instance (or \textit{spot}) when well-defined events occur. We consider event \textit{spotting} and event \textit{detection} as equivalent terminology, and use the latter moving forward.


\section{Golf Swing Sequencing}

Drawing inspiration from Giancola \textit{et al.}~\cite{giancola2018soccernet}, we describe the task of golf swing sequencing as the detection of events in \textit{trimmed} videos containing a single golf swing. The reasoning behind using trimmed golf swing videos is three-fold: 
\begin{enumerate}
    \item We speculate that the most compelling use-case for detecting golf swing events is to obtain instant biomechanical feedback in the field, as opposed to localizing golf swings in broadcast video. Although GolfDB contains the necessary data to perform spatio-temporal localization of full golf swings in untrimmed video, we consider this a separate task and a potential avenue for future research. 
    \item Golfers or golf instructors who wish to view a golf swing sequence in the field can simply record a constrained video of a single golf swing on a mobile device, ensuring that the subject is centered in the frame. This eliminates the need for spatio-temporal localization.
    \item A video sample containing a single golf swing instance will consist of a specific number of events occurring in a specific order. This information can be leveraged to improve detection performance.
\end{enumerate}

\noindent\textbf{Golf Swing Events.} In \cite{giancola2018soccernet}, soccer events were resolved at a one-second resolution. In contrast, golf swing events can be localized to a single frame using strict event definitions. Although various golf swing events have been proposed in the literature~\cite{kwon2013validity}, we define the eight contiguous events comprising a golf swing sequence as follows:
\begin{enumerate}
    \item \textit{Address (A).} The moment just before the takeaway begins, \textit{i.e.}, the frame before movement in the backswing is noticeable.
    \item \textit{Toe-up (TU).} Shaft parallel with the ground during the backswing. 
    \item \textit{Mid-backswing (MB).} Arm parallel with the ground during the backswing.
    \item \textit{Top (T).} The moment the golf club changes directions at the transition from backswing to downswing.
    \item \textit{Mid-downswing (MD).} Arm parallel with the ground during the downswing.
    \item \textit{Impact (I).} The moment the clubhead touches the golf ball.
    \item \textit{Mid-follow-through (MFT).} Shaft parallel with the ground during the follow-through. 
    \item \textit{Finish (F).} The moment just before the golfer's final pose is relaxed.
\end{enumerate}
The above definitions do not always isolate a single frame. For example, a golfer may not hold a finishing pose at all, making the selection of the \textit{Finish} event frame subjective. These issues are discussed further in the next section.

\section{GolfDB}
In this section we introduce GolfDB, a high-quality video dataset created for general recognition applications in golf, and specifically for the novel task of golf swing sequencing.  Comprising 1400 golf swing video samples and over 390k frames, GolfDB is relatively large for a specific domain. To our best knowledge, GolfDB is the first substantial dataset dedicated to computer vision applications in the sport of golf.

\begin{figure*}[t]
    \centering
    \includegraphics[width=1.0\linewidth, trim={0 0 0 1.4cm},clip]{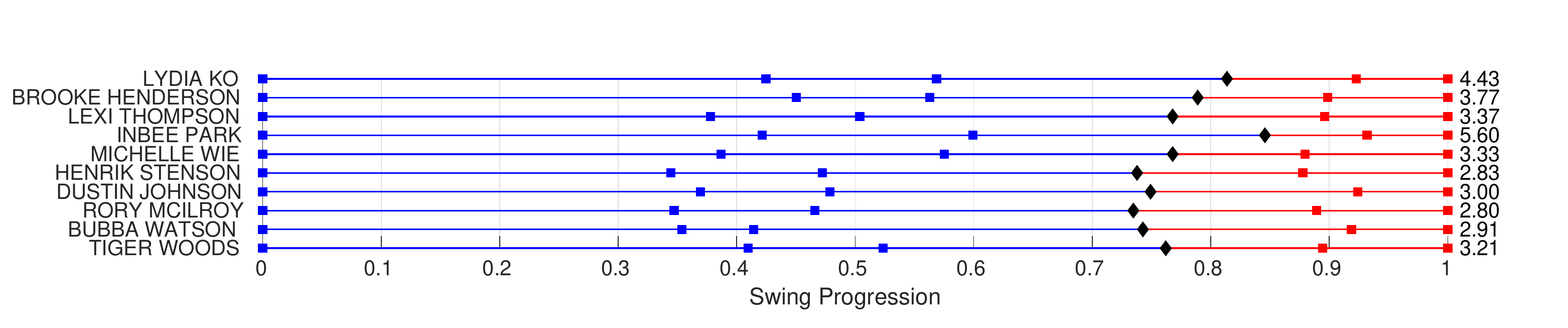}
    \vspace{-20pt}
    \caption{Average event timings from \textit{Address} to \textit{Impact} for 5 female (top 5) and male (bottom 5) professional golfers using a driver or fairway wood. Event timings normalized from \textit{Address} to \textit{Impact}. The diamond indicates the \textit{Top} event. Blue represents the backswing, red represents the downswing. Average golf swing tempos (backswing duration$/$downswing duration) shown on the right.}
    \label{fig:tempo}
    \vspace{-5pt}
\end{figure*}

\subsection{Video Collection}
A collection of 580 YouTube videos containing real-time and slow-motion golf swings was manually compiled. For the task of golf swing sequencing, it is important that the shaft remains visible at all times. To alleviate obscurities caused by motion blur, only high quality videos were considered. The YouTube videos primarily consist of professional golfers from the PGA, LPGA and Champions Tours, totalling 248 individuals with diverse golf swings. The videos were captured from a variety of camera angles, and a variety of locations on various golf courses, including the driving range, tee boxes, fairways, and sand traps. The significant variance in overall appearance, taking into consideration the different players, clubs, views, lighting conditions, surroundings, and native frame-rates, benefits the generalization ability of computer vision models trained on this dataset. The YouTube videos were sampled at 30 fps and 720p resolution. 

\subsection{Annotation}
\label{sec:annotation}
A total of 1400 trimmed golf swing video samples were extracted from the collection of YouTube videos using an in-house MATLAB code that was distributed to four annotators. For each YouTube video, the annotators were asked to identify full golf swings (\textit{i.e.}, excluding pitch shots, chips, and putts) and label 10 frames for each: the start of the sample, eight golf swing events, and the end of the sample. The number of frames between the start of the sample and \textit{Address}, and similarly, between \textit{Finish} and the end of the sample, was naturally random, and the beginning of samples occasionally included practice swings. Depending on the native frame-rate of the sample, it was not always possible to label events precisely. For example, in real-time samples with a native frame-rate of 30 fps, it was rare that the precise moment of impact was captured. The annotators were advised to chose the frame closest to when the event occurred at their own discretion.

\begin{figure}
    \vspace{-5pt}
    \centering
    \begin{subfigure}{0.49\linewidth}
        \includegraphics[width=\textwidth,trim={3.2cm 3.5cm 3cm 3.5cm},clip]{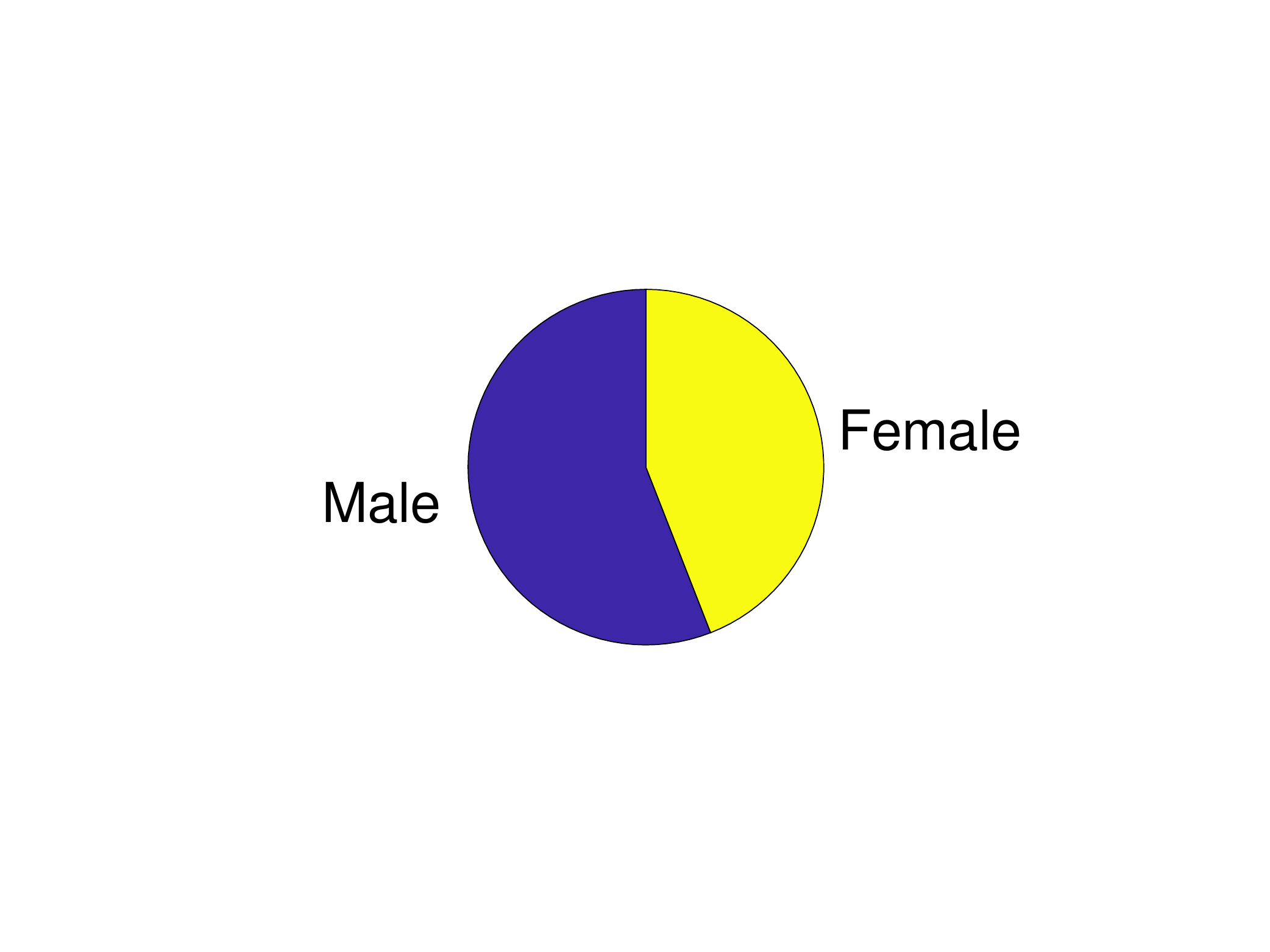}
    \end{subfigure}
    \begin{subfigure}{0.49\linewidth}
        \includegraphics[width=\textwidth,trim={3.2cm 3.5cm 3cm 3.5cm},clip]{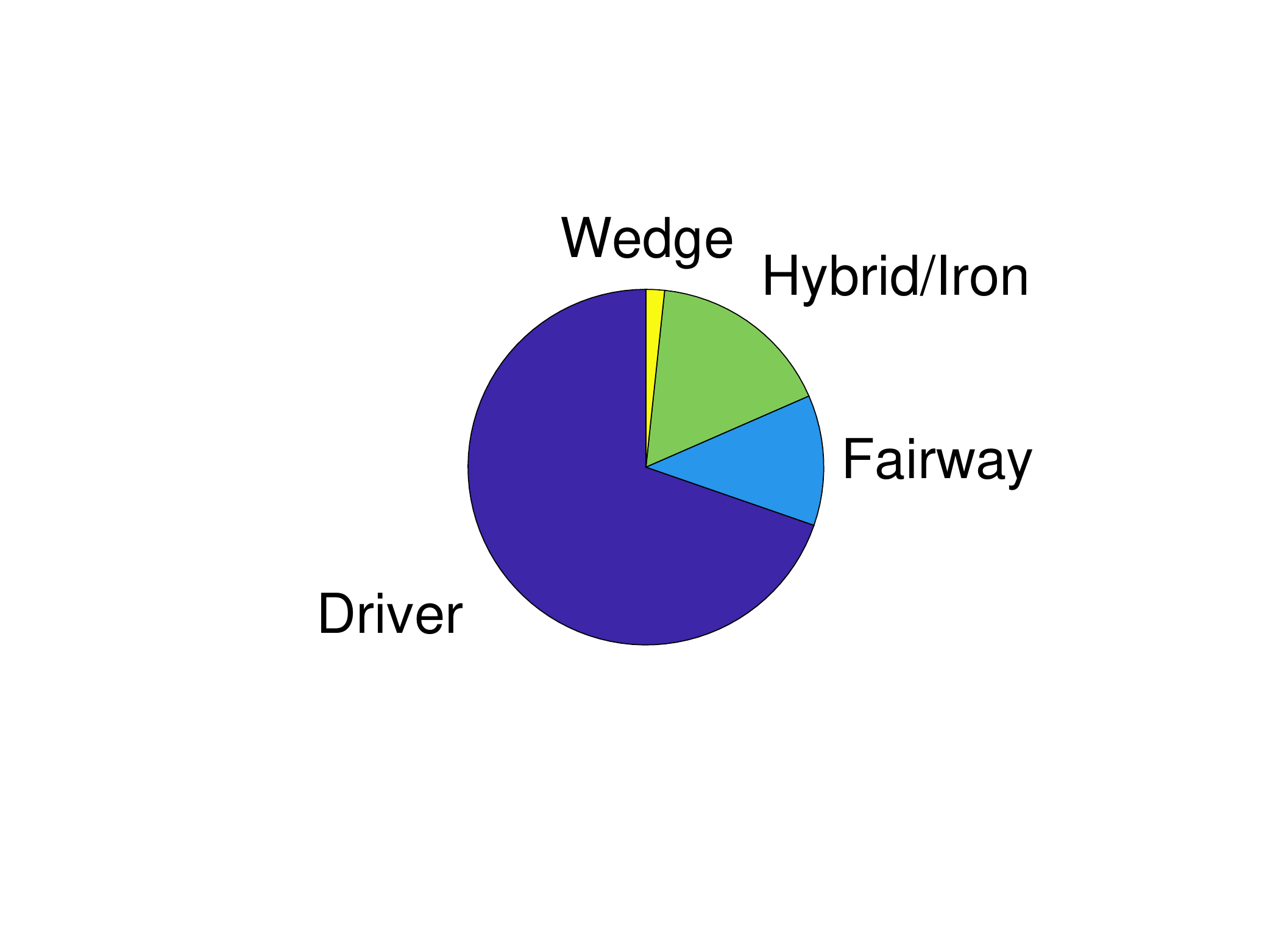}
    \end{subfigure}
    \begin{subfigure}{0.49\linewidth}
        \includegraphics[width=\textwidth,trim={3.2cm 3.5cm 3cm 3.5cm},clip]{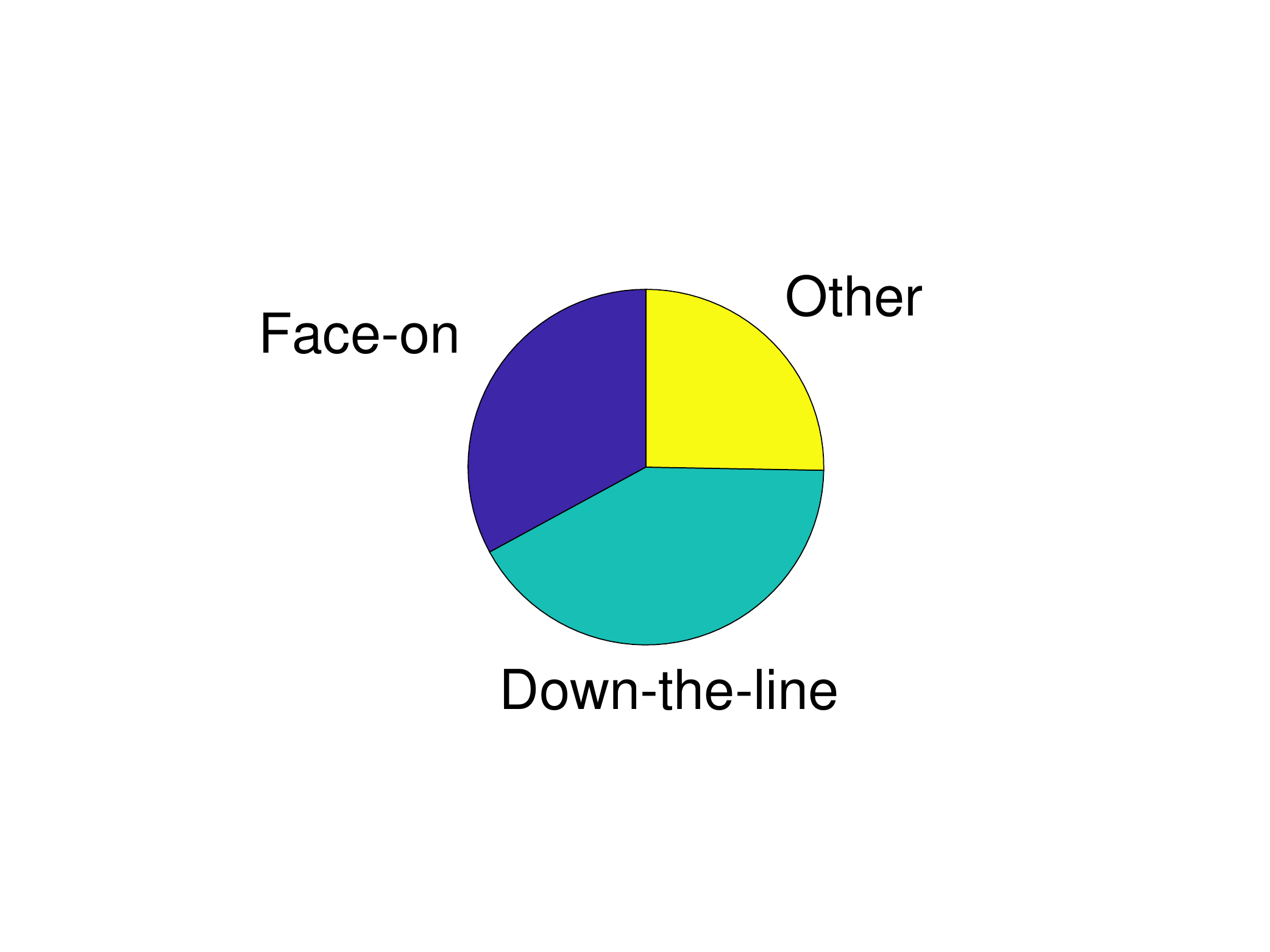}
    \end{subfigure}
    \begin{subfigure}{0.49\linewidth}
        \includegraphics[width=\textwidth,trim={3.2cm 3.5cm 3cm 3.5cm},clip]{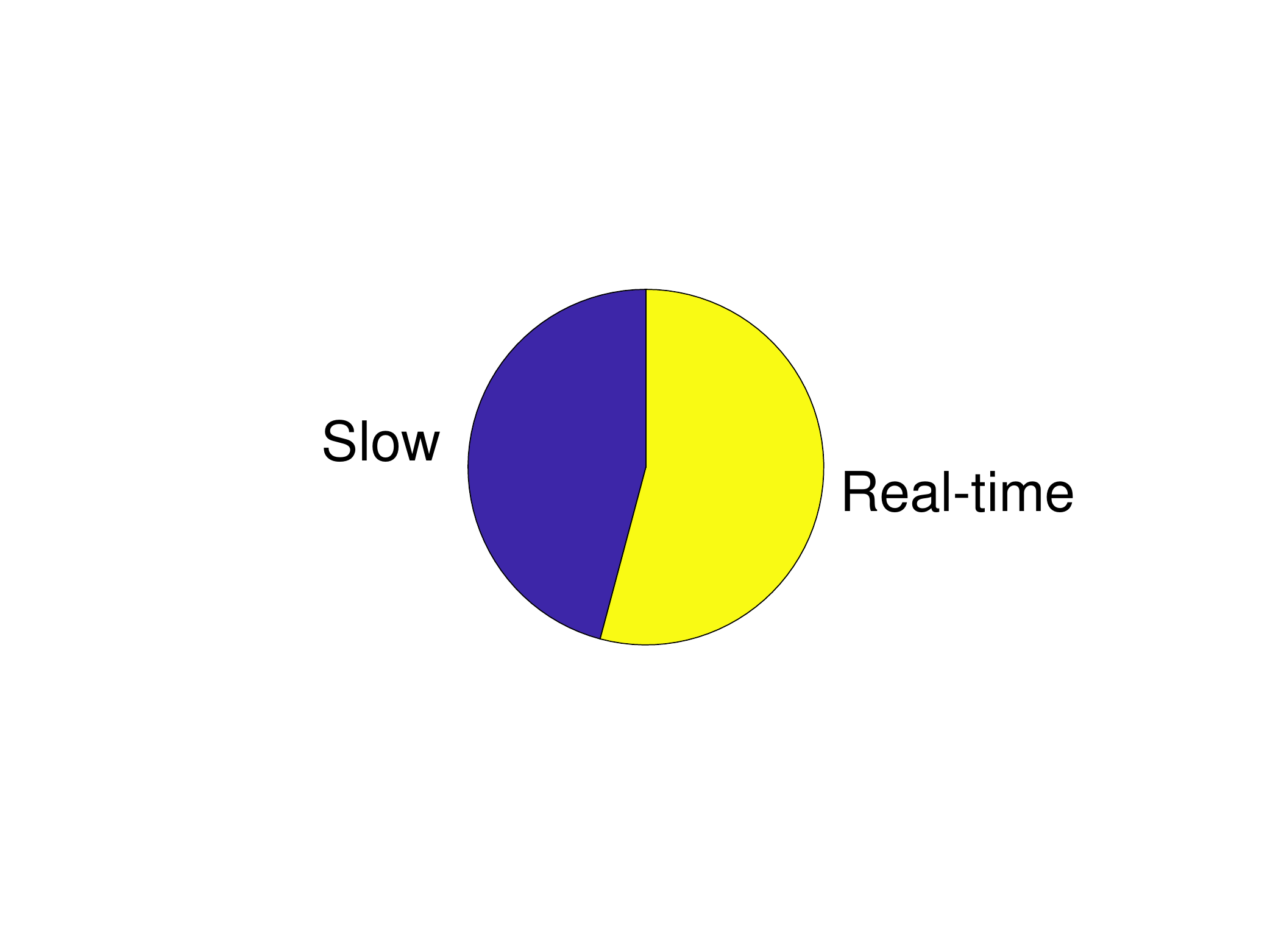}
    \end{subfigure}
    \caption{Distribution of GolfDB. The total number of frames in real-time and slow-motion samples was roughly equal ($\approx$195k each). The event densities for real-time and slow-motion samples were $3.072\times10^{-2}$ and $2.628\times10^{-2}$ events$/$frame, respectively.}
    \label{fig:stats}
    \vspace{-10pt}
\end{figure}

Besides labeling events, the annotators were asked to draw bounding boxes, enter the club and view type, and indicate whether the sample was in real-time or slow-motion (considering a playback speed of 30 fps). The bounding boxes were drawn to include the clubhead and golf ball through the full duration of the swing. Player names were extracted from the video titles, and sex was determined by cross-referencing the player name with information available online. The annotators were briefed on domain-specific knowledge before the annotation process, and the dataset was verified for quality by an experienced golfer. Fig.~\ref{fig:tempo} provides the average timing of events from \textit{Address} to \textit{Impact} for five male and female golfers using a driver or fairway wood, and Fig.~\ref{fig:stats} illustrates the distribution of the dataset.

\subsection{Evaluation Metric and Experimental Protocol}
\label{sec:protocol}
In a similar fashion to Giancola \textit{et al.}~\cite{giancola2018soccernet}, we introduce a tolerance $\delta$ on the number of frames within which an event is considered to be correctly detected. In the real-time samples, if an event was thought to occur between two frames, it was at the discretion of the annotator to select the event frame. Given the inherent variability of the annotator's selection, we consider a tolerance $\delta = 1$ for real-time videos sampled at 30 fps. For the slow-motion videos, the tolerance should be scaled based on the native frame-rate, but the native-frame rates are unknown. We therefore propose a sample-dependent tolerance based on the number of frames between \textit{Address} and \textit{Impact}. This value was approximately 30 frames on average for the real-time videos, matching the frame-rate of 30 fps (\textit{i.e.}, the average duration of a golf swing from \textit{Address} to \textit{Impact} is roughly 1s). Thus, we define the sample-dependent tolerance as
\begin{equation}
    \delta = \max(\nint*{\frac{n}{f}}, 1)
\end{equation}
where $n$ is the number of frames from \textit{Address} to \textit{Impact}, $f$ is the sampling frequency, and $\nint{x}$ rounds $x$ to the nearest integer.

Drawing inspiration from the field of human pose estimation, we introduce the PCE evaluation metric as the ``Percentage of Correct Events'' within tolerance. PCE is the temporal equivalent to the popular PCKh metric used in human pose estimation~\cite{andriluka14cvpr}, which scales a spatial detection tolerance using head segment length. For the experimental protocol, four random splits were generated for cross-validation, ensuring that all samples from the same YouTube video were placed in the same split. PCE averaged over the 4 splits is used as the primary evaluation metric.


\section{SwingNet: A Swing Sequencing Network}
In this section, we describe SwingNet, a network architecture designed specifically for the task of golf swing sequencing, but generalizable to the swings present in various sports, such as baseball, tennis and cricket. Additionally, the implementation details are discussed. 

\subsection{Network Architecture}

\begin{figure*}[t]
    \centering
    \includegraphics[width=1.0\linewidth]{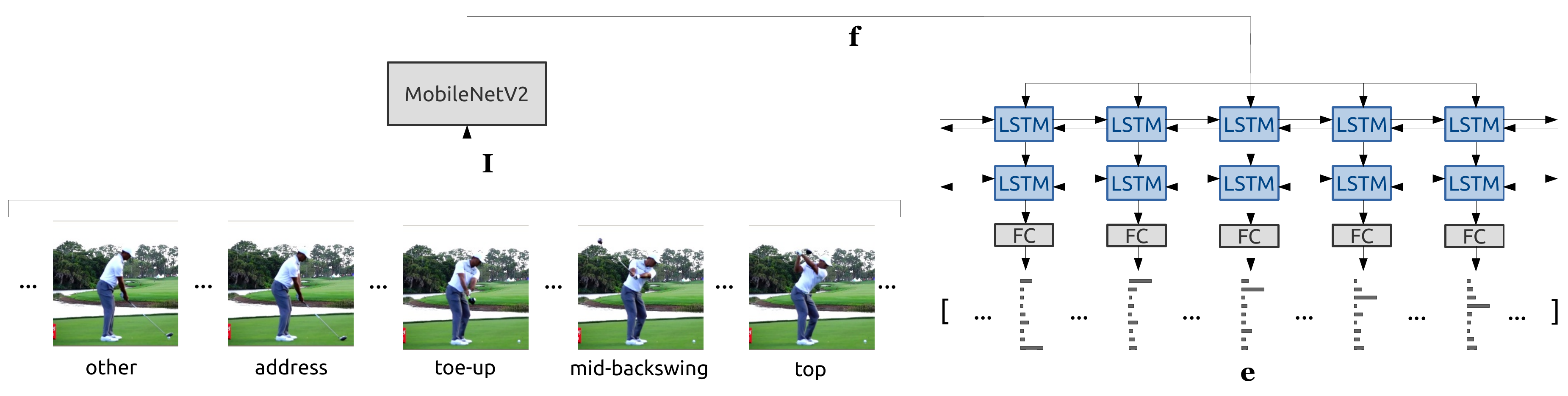}
    \caption{The network architecture of SwingNet, a deep hybrid convolutional and recurrent network for swing sequencing. In an end-to-end architectural framework, SwingNet maps a sequence of RGB images $\mathbf{I}$ to a corresponding sequence of event probabilities $\mathbf{e}$. The sequence of feature vectors $\mathbf{f}$ generated by MobileNetV2 are input to a bidirectional LSTM. At each frame $t$, the LSTM output is fed into a fully-connected layer, and a softmax is applied to obtain the event probabilities.}
    \label{fig:arch}
    \vspace{-10pt}
\end{figure*}

MobileNetV2 is a CNN based on an inverted residual structure and makes liberal use of lightweight depthwise convolutions~\cite{howard2017mobilenets}. As such, it is well suited for mobile vision applications. Furthermore, MobileNetV2 includes a ``width multiplier'' that scales the number of channels in each layer, providing a convenient trade-off for network complexity and speed. For the task of image classification, it runs at 75ms per frame on a single core of the Google Pixel using an input size of 224$\times$224 and width multiplier of 1~\cite{Sandler_2018_CVPR}.  Placing emphasis on mobile deployment, we employ MobileNetV2~\cite{Sandler_2018_CVPR} as the backbone CNN in an end-to-end network architecture that maps a sequence of $d \times d$ RGB images  $\mathbf{I} = (\mathbf{I_1}, \mathbf{I_2}, ..., \mathbf{I_T} : \mathbf{I_t} \in \mathbb{R}^{d \times d \times 3})$ to a corresponding sequence of event probabilities $\mathbf{e} = (\mathbf{e_1}, \mathbf{e_2}, ..., \mathbf{e_T} : \mathbf{e_t} \in \mathbb{R}^C)$, where $T$ is the sequence length and $C$ is the number of event classes. For the task of golf swing sequencing, there are 9 event classes: eight golf swing events and one \textit{No-event} class. 

Detecting golf swing events using a single frame would likely be a difficult task. Precisely identifying \textit{Address} requires knowledge of when the \textit{actual} golf swing commences. Without this contextual information, \textit{Address} may be falsely detected during the pre-shot routine, which often includes full or partial practice swings, and frequent clubhead waggling. In a similar manner, precisely identifying \textit{Top} is generally not possible using a single frame, based on its event definition. Moreover, \textit{Mid-backswing} and \textit{Mid-downswing} are relatively similar in appearance. For these reasons, temporal context is likely a critical component in the task of golf swing sequencing. To capture temporal information, the sequence of feature vectors $\mathbf{f} = (\mathbf{f_1}, \mathbf{f_2}, ..., \mathbf{f_T} : \mathbf{f_t} \in \mathbb{R}^{1280})$ obtained by applying global average pooling to the final feature map in MobileNetV2 is used as input to an $N$-layer bidirectional LSTM~\cite{hochreiter1997long} with $H$ hidden units in each layer. At each frame $t$, the $H$-dimensional output of the LSTM is fed through a final fully-connected layer, and a softmax is applied to obtain the class probabilities $\mathbf{e}$. The weights of the fully-connected layer are shared across frames. The overall architecture is illustrated in Fig.~\ref{fig:arch}. In Section~\ref{sec:hp_search}, an ablation study is performed to identify a suitable model configuration.

\subsection{Implementation Details}
\label{sec:implementation}

The frames were cropped using the golf swing bounding boxes. They were then resized using bilinear interpolation such that the longest dimension was equal to $d$, and padded using the ImageNet~\cite{imagenet_cvpr09} pixel means to reach an input size of $d \times d$. Finally, all frames were normalized by subtracting the ImageNet means and dividing by the ImageNet standard deviation.

The network was implemented using PyTorch version 1.0. Each convolutional layer in MobileNetV2\footnote{PyTorch implementation of MobileNetV2 available at \url{https://github.com/tonylins/pytorch-mobilenet-v2}} is followed by batch normalization~\cite{ioffe2015batch}. Unique to MobileNetV2, ReLU is used after batch normalization everywhere except for the final convolution in the inverted residual modules~\cite{Sandler_2018_CVPR}, where no non-linearity is used. The running batch norm parameters were updated using a momentum of 0.1. The MobileNetV2 backbone was initialized with weights pre-trained on ImageNet, and the weights in the fully-connected layer were initialized following Xavier Initialization~\cite{glorot2010understanding}. Given the significant class imbalance between the golf swing events and the \textit{No-event} class (roughly 35:1), a weighted cross-entropy loss was used, where the golf swing events were each given a weight of 1, and the \textit{No-event} class was given a weight of 0.1.

Training samples were drawn from the dataset by randomly selecting a start frame and, if the number of frames remaining in the sample was less than $T$, the sample was looped to the beginning. Randomly selecting the start frame serves as a form of data augmentation and is commonly used in action recognition applications~\cite{carreira2017quo, mcnally2019starnet}.  Other forms of data augmentation used included random horizontal flipping, and slight random affine transformations ($-5^\circ$ to $+5^\circ$ shear and rotation). The intent of the horizontal flipping was to even out the imbalance between left- and right-handed golfers, and the intent of the affine transformations was to capture variations in camera angle. To enable training on batches of image sequences, multiple sequences of length $T$ were concatenated along the channel dimension before being input to the network. The output features $\mathbf{f}$ were reshaped to (batch size, $T$, 1280) before being passed to the LSTM. Various batch sizes and sequence lengths $T$ were explored in the experiments; they are explicitly declared in Section~\ref{sec:exp}. In all training configurations, the Adam optimizer~\cite{kingma2014adam} was used with an initial learning rate of 0.001. The number of training iterations and learning rate schedules are discussed in Section~\ref{sec:exp}. All training was performed on a single NVIDIA Titan Xp GPU. 

At test time, a sliding window approach is used over the full-length golf swing video samples. The sliding window has a size $T$ and, to minimize the computational load, no overlap was used. Knowing that each sample contains exactly eight events, event frames were selected using the maximum confidence for each event class. The exploration of alternative selection criteria that leverages event order is encouraged in future research. 

\section{Experiments}
\label{sec:exp}

In this section, an extensive ablation study is performed to determine suitable hyper-parameters. Following the ablation study, a final baseline SwingNet is proposed and evaluated.

\subsection{Ablation Study}
\label{sec:hp_search}

\begin{table*}[t]
\begin{center}
\begin{tabular}{|c|c|c|c|c|c|c|c|c|c|c|}
\hline
Config. & \makecell{Input\\Size\\($d$)} & \makecell{Seq.\\Length\\($T$)} & \makecell{Batch\\Size} & \makecell{ImageNet\\Weights} & \makecell{LSTM\\Layers\\($N$)} & \makecell{LSTM\\Hidden\\($H$)} & Bidirect. & \makecell{Params\\($10^6$)} & \makecell{FLOPs\\($10^9$)} & \makecell{PCE\\(10k iter.)} \\
\hline\hline
0 & \textbf{224} & 32 & 6 & \textbf{Yes} & \textbf{2} & \textbf{128} & \textbf{Yes} & 4.07 & 10.32 & 66.8\\
1 & 224 & 32 & 6 & \textbf{No} & 2 & 128 & Yes & 4.07 & 10.32 & 1.5\\
2 & 224 & 32 & 6 & Yes & 2 & 128 & \textbf{No} & 3.08 & 10.26 & 54.7\\
3 & \textbf{192} & 32 & 6 & Yes & 2 & 128 & Yes & 4.07 & 7.62 & 45.7\\
4 & \textbf{160} & \textbf{32} & \textbf{6} & Yes & 2 & 128 & Yes & 4.07 & 5.33 & 62.4\\
5 & \textbf{128} & 32 & 6 & Yes & 2 & 128 & Yes & 4.07 & 3.45 & 57.7\\
6 & 160 & \textbf{64} & 6 & Yes & 2 & 128 & Yes & 4.07 & 10.65 & 71.1\\
7 & 160 & 32 & \textbf{12} & Yes & 2 & 128 & Yes & 4.07 & 5.33 & 70.1\\
8 & 224 & 32 & 6 & Yes & \textbf{1} & 128 & Yes & 3.67 & 10.39 & 69.4\\
9 & 224 & 32 & 6 & Yes & 2 & \textbf{64} & Yes & 3.01 & 10.26 & 66.9\\
10 & 224 & 32 & 6 & Yes & 2 & \textbf{256} & Yes & 6.96 & 10.51 & 69.3\\
\hline 
\end{tabular}
\end{center}
\vspace{-10pt}
\caption{Hyper-parameter search for the proposed SwingNet. Each configuration was trained for 10k iterations on the first split of GolfDB, providing a proxy for final performance. Parameters used
in comparison are in \textbf{bold}.}
\vspace{-10pt}
\label{tab:ablation}
\end{table*}

The hyper-parameters of interest are the input size ($d$), sequence length ($T$), batch size, number of LSTM layers ($N$), and number of hidden units in each LSTM layer ($H$). It was also of interest to see whether initializing with pre-trained ImageNet weights was advantageous as opposed to training from scratch, and if LSTM bidirectionality had an impact. The goal was to identify a computationally efficient configuration to maximize performance with limited compute resources. Normally, MobileNetV2's width multiplier would be a key hyper-parameter to include in the ablation study; however, for reasons to be discussed, using a width multiplier other than 1 was not feasible. Table~\ref{tab:ablation} provides the PCEs of 11 model configurations, along with the number of parameters and floating point operations (FLOPs) for each. Each configuration was trained for 10k iterations on the first split of GolfDB, providing a proxy for overall performance. For the ablation study, no affine transformations were used, and the learning rate was held constant at 0.001. Hyper-parameters used in comparison are shown in bold. 

Remarkably, it was found that the model did not train at all unless the pre-trained ImageNet weights were used. Recent research from Facebook AI~\cite{he2018rethinking} found that initializing with pre-trained ImageNet weights did not improve performance for object detection on the COCO dataset~\cite{lin2014microsoft}. COCO is a large-scale dataset that likely has a similar distribution to ImageNet. Thus, we speculate that using pre-trained weights is critical in domain-specific tasks, where the variation in overall appearance is minimal. Because the pre-trained weights were only available for a width multiplier of 1, adjusting the width multiplier was not feasible.

Another interesting finding was that the correlation between input size and performance did not behave as expected. With increasing input size, one would expect a monotonically increasing PCE. However, it was found that that input sizes of 160 and 128 outperformed 192 by a large margin, and the PCE using an input size of 160 was only 4.4 points worse than the 224. 


The sequence length $T$ had a significant impact on performance, supporting the importance of temporal context. Additionally, increasing the batch size improved performance dramatically. Still, it is difficult to fairly assess these hyper-parameters as they may have had a significant impact on convergence speed. Regarding the LSTM, bidirectionality led to a 12.1 point improvement in PCE. The single-layer LSTM outperformed the two-layer LSTM, and 256 hidden units performed better than 64 and 128.

\subsection{Frozen Layers}
\begin{figure}[t]
    \centering
    \includegraphics[width=1.0\linewidth]{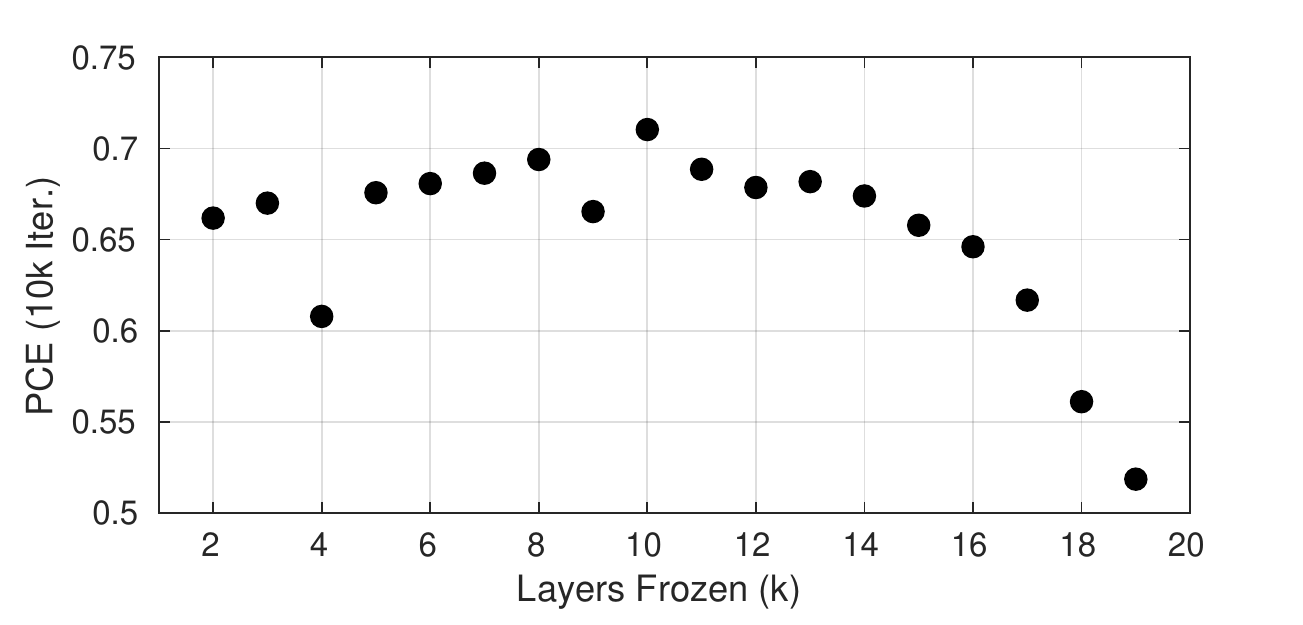}
    \caption{PCE after 10k iterations using configuration 4 from Table~\ref{tab:ablation} and freezing the ImageNet weights in the first $k$ layers. Freezing the first 10 layers provided optimal results.}
    \label{fig:frozen}
    \vspace{-10pt}
\end{figure}

The results of the ablation study revealed that increasing the sequence length and batch size provided large performance gains. On a single GPU, the sequence length and batch size are severely limited. Knowing that the model relies heavily on pre-trained ImageNet weights, we hypothesized that some of the ImageNet weights could be frozen without a significant loss in performance. Thus, we experimented with freezing layers in MobileNetV2 prior to training to create space in GPU memory for larger sequence lengths and batch sizes. Fig.~\ref{fig:frozen} plots the results of freezing the first $k$ layers using configuration 4 from Table~\ref{tab:ablation}. Despite a few outliers, the PCE increased until $k = 10$ and then began to decrease. We leverage this finding in the next section to train a baseline SwingNet using a larger sequence length and batch size.

\subsection{Baseline SwingNet}

\begin{table*}[t]

\begin{center}
\begin{tabular}{|l|c|c|c|c|c|c|c|c|c|}
\hline
Model & A & TU & MB & T & MD & I & MFT & F & PCE\\
\hline\hline
SwingNet-160 (slow-motion) & 23.5 & 80.7 & 84.7 & 75.7 & 97.8 & 98.3 & 98.0 & 21.5 & 72.5 \\
SwingNet-160 (real-time) & 38.7 & 87.2 & 92.1 & 90.8 & 98.3 & 98.4 & 97.2 & 30.7 & 79.2 \\
\hline
SwingNet-160 & 31.7 & 84.2 & 88.7 & 83.9 & 98.1 & 98.4 & 97.6 & 26.5 & 76.1 \\
\hline 
\end{tabular}
\end{center}
\vspace{-5pt}
\caption{Event-wise and overall PCE averaged over the 4 splits for the proposed baseline SwingNet. Configuration: bidirectional LSTM, $d=160$, $T=64$, $L=1$, $N=256$, $k=10$. This configuration has $5.38 \times 10^6$ parameters and $10.92 \times 10^9$ FLOPs.}
\vspace{-5pt}
\label{tab:baseline}
\end{table*}

\begin{table}[t]
\begin{center}
\begin{tabular}{|c|c|c|c|}
\hline
\makecell{Seq. Length ($T$)} & FLOPs ($10^9$) & CPU (ms)$^*$ & PCE\\
\hline\hline
64 & 10.92 & 10.6 & 76.2\\
32 & 5.41 & 10.8 & 74.0\\
16 & 2.70 & 11.5 & 71.0\\
8 & 1.35 & 12.0 & 66.0\\
4 & 0.68 & 13.8 & 63.1\\
\hline
\end{tabular}
\end{center}
\vspace{-5pt}
\caption{SwingNet-160 performance and CPU runtime on GolfDB split 1 using various sequence lengths at test time. $^*$Effective processing time for a single frame, excluding I/O, using an Intel i7-8700K processor.}
\label{tab:seq_length}
\vspace{-10pt}
\end{table}

\begin{figure*}[b!]
    \centering
    \includegraphics[width=1.0\linewidth]{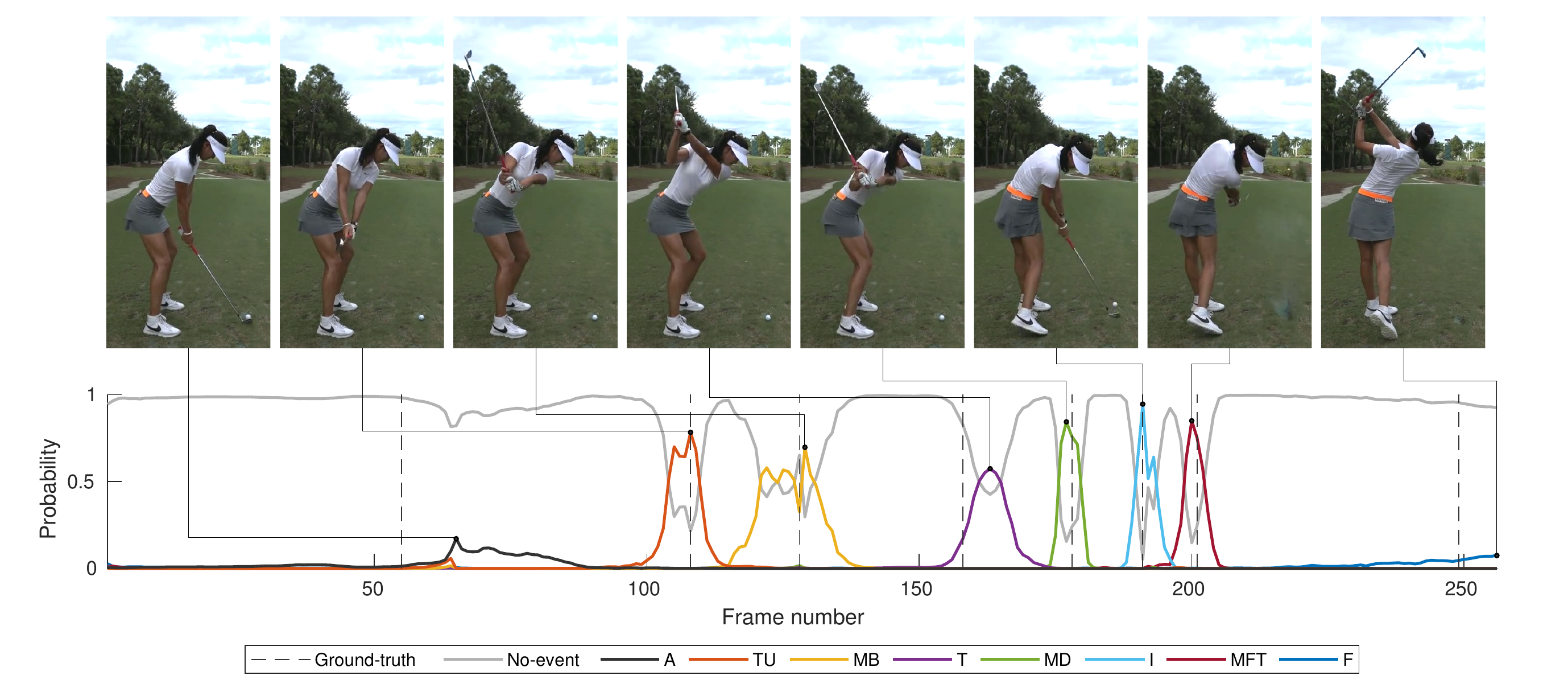}
    \vspace{-20pt}
    \caption{Using SwingNet to infer event probabilities in the slow-motion golf swing of LPGA Tour player Michelle Wie. Six out of eight events were correctly detected within a 5-frame tolerance. \textit{Address} and \textit{Finish} were missed by 10 and 7 frames, respectively. Best viewed in color. Video available at \url{https://youtu.be/QlKodM7RhH4?t=36}. }
    \label{fig:inference}
\end{figure*}

The results in Table~\ref{tab:ablation} suggest that an input size of 160 is more cost effective than an input size of 224; the latter requires double the computation for a relative increase in PCE of just 7\%. Taking this into consideration, as well as the comparative results of the other hyper-parameters, we propose a baseline SwingNet with an input size of 160, sequence length of 64, and a single-layer bidirectional LSTM with 256 hidden units. By initializing with pre-trained ImageNet weights and freezing 10 layers, the model can be trained using a batch size of 24 on a single GPU with 12GB memory. With this relatively large batch size, the model converges faster and does not need to be trained for as many iterations. Thus, the baseline SwingNet was trained for 7k iterations on each split of GolfDB, and the learning rate was reduced by an order of magnitude after 5k iterations. Affine transformations were used to augment the training data (see Section~\ref{sec:implementation}). 


Table~\ref{tab:baseline} provides the event-wise and overall PCE averaged over the 4 splits of GolfDB. Overall, SwingNet correctly detects events at a rate of 76.1\%. As expected, relatively poor detection rates were observed for the \textit{Address} and \textit{Finish} events. This was likely caused by the compounding factors of subjective labeling and the inherent difficulty associated with precisely localizing these events temporally. These factors may have also played a role in detecting the \textit{Top} event, which was detected in real-time videos more consistently than in slow-motion videos; in slow-motion, the exact frame where the club changes directions is difficult to interpret because the transition is more fluid. Moreover, the detection rate was generally lower in slow-motion videos for the backswing events. This was likely due to the fact that the backswing is much slower than the downswing, so there are more frames in the backswing that are similar in appearance to the ground-truth event frames. 

\textit{Impact} was the event detected the most proficiently. This result is intuitive because the model simply needs to detect when the clubhead is nearest the golf ball. Interpreting when the arm or shaft is parallel with the ground, which is required for events like \textit{Toe-up} and \textit{Mid-backwing}, requires more abstract intuition. Disregarding the \textit{Address} and \textit{Finish} events, the overall PCE was 91.8\%. Fig.~\ref{fig:inference} illustrates the inferred event probabilities for a slow-motion swing. Within a 5-frame tolerance, SwingNet correctly detected all but the \textit{Address} and \textit{Finish} events, which were off by 10 and 7 frames from their respective ground-truth frames. 

Table~\ref{tab:seq_length} demonstrates that, after being trained using a sequence length of 64, smaller sequence lengths can be used at test time with only a modest decrease in performance. This has implications to mobile deployment, where smaller sequence lengths may be leveraged to reduce the memory requirements of the network. 

\section{Conclusion}
This paper introduced the task of golf swing sequencing as the detection of key events in trimmed golf swing videos. The purpose of golf swing sequencing is to facilitate golf swing analysis by providing instant feedback in the field through the automatic extraction of key frames on mobile devices. To this end, a golf swing video database (GolfDB) was established to support the task of golf swing sequencing. Advocating mobile deployment, we introduce SwingNet, a lightweight baseline network with a deep hybrid convolutional and recurrent network architecture. Experimental results showed that SwingNet detected eight golf swing events at an average rate of 76.1\%, and six out of eight events at a rate of 91.8\%. Besides event labels, GolfDB also contains annotations for golf swing bounding boxes, player name and sex, club type, and view type. We provide this data with the intent of promoting future computer vision research in golf, such as the spatio-temporal localization of golf swings in untrimmed broadcast video, and view type recognition.

\noindent\textbf{Acknowledgments.} We acknowledge the NVIDIA GPU Grant Program, and financial support from the Canada Research Chairs program and the Natural Sciences and Engineering Research Council of Canada (NSERC).

{\small
\bibliographystyle{ieee}
\bibliography{GolfDB}
}

\end{document}